\documentclass[10pt,twocolumn,letterpaper]{article}

\usepackage{iccv}
\usepackage{times}
\usepackage{epsfig}
\usepackage{graphicx}
\usepackage{amsmath}
\usepackage{amssymb}
\usepackage{booktabs}  
\usepackage{mathrsfs}  
\usepackage{cuted}
\usepackage{capt-of}

\usepackage[pagebackref=true,breaklinks=true,letterpaper=true,colorlinks,bookmarks=false]{hyperref}

\iccvfinalcopy % *** Uncomment this line for the final submission

\def\iccvPaperID{1938} % *** Enter the ICCV Paper ID here

\ificcvfinal\fi

\usepackage{color}
\definecolor{green}{rgb}{0, 0.5, 0}
\definecolor{orange}{rgb}{0.8, 0.6, 0.2}
\definecolor{red}{rgb}{1.0, 0.0, 0.0}
\definecolor{teal}{rgb}{0.0, 0.4, 0.4}
\definecolor{purple}{rgb}{0.65,0,0.65}
\definecolor{saffron}{rgb}{0.95,0.75,0.2}
\definecolor{turquoise}{rgb}{0.0,0.5,0.5}
\definecolor{black}{rgb}{0.0, 0.0, 0.0}
\definecolor{gray}{rgb}{0.5, 0.5, 0.5}

\newcommand{\rz}[1]{{\color{black}#1}}

\newcommand{\mypara}[1]{\noindent \textit{#1}.}

\begin{document}

%%%%%%%%% TITLE
\title{D-Fusion: Discriminated Fusion of Texts and Styles via Stable Diffusion}
\title{DS-Fusion: Artistic Typography via Discriminated and Stylized Diffusion}

\author{Maham Tanveer$^1$, Yizhi Wang$^1$, Ali Mahdavi-Amiri$^1$, Hao Zhang$^1$ \\ $^1$Simon Fraser University}

\twocolumn[{
% \renewcommand
% \twocolumn[1][]{#1}%
\maketitle
\begin{center}
    \centering
    % \captionsetup{type=figure}
    \includegraphics[width=1\textwidth]{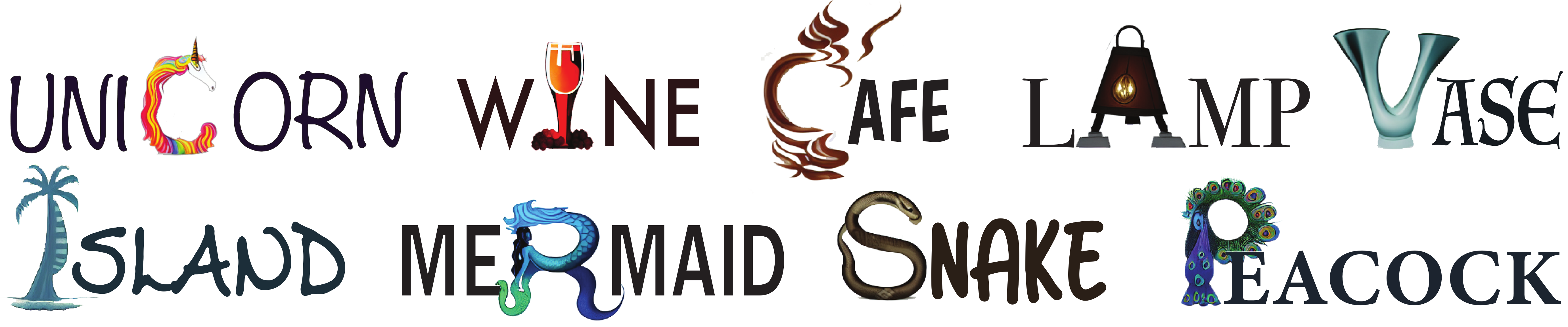}
    \captionof{figure}{\rz{Single-letter artistic typographies generated {\em fully automatically\/} by our network. Each result is produced from a prompt consisting of a word, e.g., ``unicorn", and a  letter in the word, e.g., ``C", to be stylized based on the semantics of the word and the input font.}}
    \label{fig:teaser}
\end{center}%
}]

% \begin{strip}\centering
% \includegraphics[width=\textwidth]{data/figure_teaser_bw.pdf}
% \captionof{figure}{Feature graphic caption.
% \label{fig:feature-graphic}}
% \end{strip}

\maketitle

\ificcvfinal\thispagestyle{empty}\fi

\begin{abstract}
We introduce a novel method to {\em automatically\/} generate an {\em artistic typography\/} by {\em stylizing\/} one 
or more letter fonts to visually convey the semantics of an input word, while ensuring that the output 
remains readable. To address an assortment of challenges with our task at hand including 
conflicting goals (artistic stylization vs.~legibility), lack of ground truth, and immense search space, our
approach utilizes large language models to bridge texts and visual images for stylization and build an
{\em unsupervised\/} generative model with a diffusion model backbone. Specifically, we employ the
denoising generator in Latent Diffusion Model (LDM), with the key addition of a CNN-based 
{\em discriminator\/} to adapt the input style onto the input text. The discriminator uses rasterized images of 
a given letter/word font as real samples and output of the denoising generator as fake samples.
Our model is coined DS-Fusion for discriminated and stylized diffusion. We showcase the quality and 
versatility of our method through numerous examples, qualitative and quantitative evaluation, as well
as ablation studies. User studies comparing to strong baselines including CLIPDraw and DALL-E 2,
as well as artist-crafted typographies, demonstrate strong performance of DS-Fusion.
\end{abstract}

\section{Introduction}
\label{sec:intro}

%\noindent
%``Art is a form of exploration, of sailing off into the unknown alone, heading for those unmarked places on the map.''
%\vspace{-5pt}
%\begin{flushright}
%--- Michael Chabon
%\end{flushright}

\begin{figure*}
\centering
  \includegraphics[width=1\textwidth]{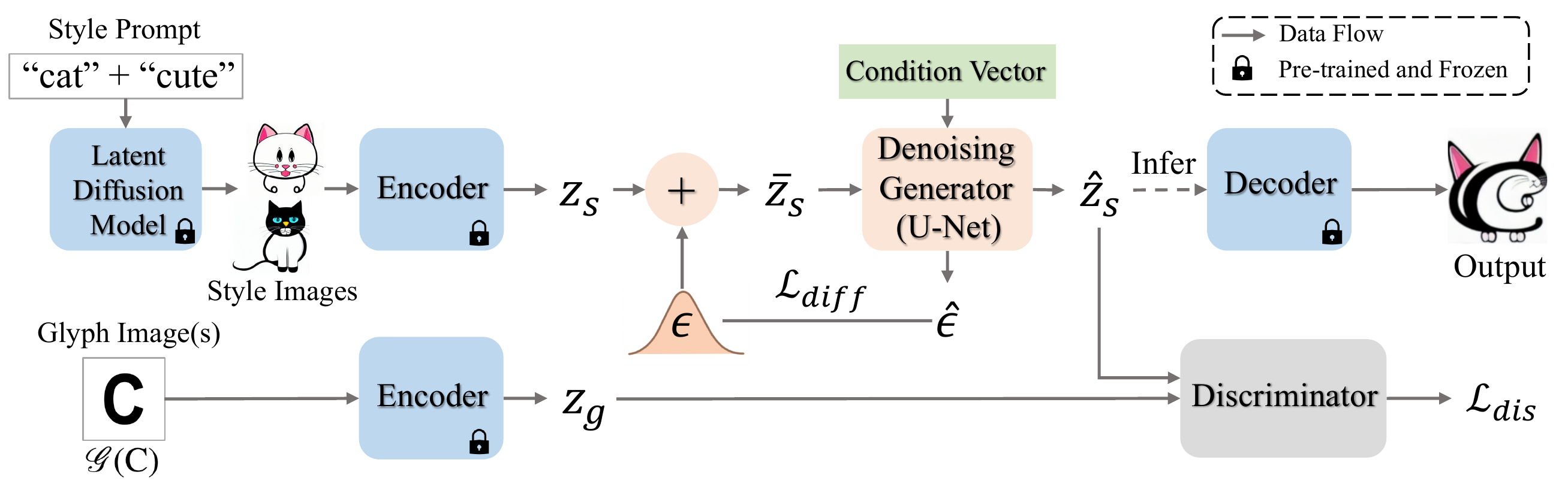}
  \caption{The pipeline of DS-Fusion, which takes as input a style prompt and a glyph image. The style images are generated according to the style word and attribute. DS-Fusion first utilizes a latent diffusion process~\cite{rombach2022high} to construct the latent space of the given style and then introduces a discriminator to blend the style into the glyph shape. The parameters of a module are pre-trained and frozen if there is an icon of a lock on the bottom right. The ``$+$'' module denotes the iterative noise injection process of diffusion models.}
  \label{fig:method_pipeline}
\end{figure*}

\rz{We explore the artistic-creative potential of automated generative processes modeled by modern neural networks. Specifically, we
are interested in the generation of {\em artistic typography\/}. According to Wikipedia, typography is the art and technique of arranging 
type to make written language legible, readable, and appealing when displayed. Artistic typography represents a style of 
typography that goes beyond the basic function of conveying information through text and seeks to create a visual impact on the reader. 
It involves using typography as a form of artistic expression and allows designers to create eye-catching typographic designs that 
express a message visually and creatively.

In this paper, we aim to automatically generate an artistic typography by {\em stylizing\/} one or more letter fonts, to visually convey 
the semantics of an input word, while ensuring that the output typography is readable; see Figure~\ref{fig:teaser} for such 
examples referred to as ``word-as-image''~\cite{berio2022strokestyles,tendulkar2019trick,zhang2017synthesizing,iluz2023word2img}.
It is an arduous task to combine semantics and text in a legible and artistic manner for several reasons. First, the goal of 
incorporating the aesthetics of a style in an abstract and creative way into a letter or word can conflict with the desire to 
maintain readability of the original word/letter. Second, what a good artistic typography is can be a subjective matter. Without
a universally accepted ``ground truth", %and a lack of large volumes of artistic creations, 
a viable learning approach will have to be {\em unsupervised\/}. Last but not least, semantics can be depicted in numerous ways. 
For instance, to indicate the presence of a lion, one can use the entire face, the tail, or the whole animal. There is a vast 
range of lion images, icons, and shapes that are accessible, making it nearly impossible to manually search, deform, and substitute them. 
While experienced artists and designers are capable of producing beautiful semantic typography, obtaining reasonable results for 
ordinary users and hobbyists without proper assistive tools are out of reach.

To address all the above challenges, we resort to recently popularized large language 
models~\cite{rombach2022high,radford2021learning} to bridge texts and visual images for stylization and build our unsupervised
generative model for artistic typography on Latent Diffusion~\cite{radford2021learning}. Specifically, we employ the {\em denoising 
generator\/} in Latent Diffusion Model,
%and their encoder, 
with the key addition of a CNN-based {\em discriminator\/} to adapt the input style 
onto the input text (Figure~\ref{fig:method_pipeline}). The discriminator uses rasterized images of a given letter/word font as real samples and output of the denoising 
generator as fake samples. To obtain images for the denoising generator to guide the letter stylization, we generate 25 style images 
from the input word, again, with Latent Diffusion. The selection of this number aims to ensure an adequate number of instances for the diffusion model to extract underlying features and attain diversity in the outputs. We fine-tune the denoising generator on 
these images using the diffusion loss based on the style images and the discriminator loss.}

Our model is coined {\em DS-Fusion\/} for discriminated and stylized diffusion. While the core idea is quite simple, it is among the first to 
integrate adversarial learning and diffusion in a single framework. By utilizing the powerful generation capabilities of diffusion models and 
employing a discriminator as a critic, we ensure that the produced artistic typography remains true to the input font. Figure~\ref{fig:teaser}
shows some results generated by DS-Fusion {\em fully automatically\/}.

We showcase the effectiveness of DS-Fusion for generating artistic typography through numerous experiments. Our approach produces visual results that demonstrate its generality and versatility in accommodating different semantics, letters, and artistic styles. We report quantitative evaluations and ablation studies to assess the contribution of individual components. Additionally, we have conducted user studies to assess the quality of our automatically generated typography results. These studies reveal that in about 42\% of the cases, our results were favored over or considered equally good as results produced by professional artists, and in close to 50\% of cases, our method outperforms the state-of-the-art alternatives, including DALL-E 2~\cite{ramesh2022hierarchical}, a strong baseline that had been trained on significantly more images than LDM. It is worth noting that we did not choose specific inputs catering to DS-Fusion for these comparisons. Instead, we performed a Google image search on ``artistic typography" and extracted a suitable subset of artist-generated results to come up with inputs for both user studies.

\section{Related work}
\label{sec:related}

\mypara{Learning fonts}
There are numerous techniques to study, design, and stylize fonts. 
%Kautz~\cite{campbell2014learning} utilized GP-LVM algorithm to learn a font manifold from  the polyline representation of glyph outlines.
Campbell and Kautz~\cite{campbell2014learning} utilized an algorithm to learn a font manifold from  the polyline representation of glyph outlines. By exploring this manifold, new fonts can be obtained, or existing fonts can be interpolated to achieve a desired effect. 
Balashova et al.~\cite{balashova2019learning} proposed an approach that uses a stroke-based geometric model for glyphs and a fitting procedure to reparametrize arbitrary fonts to the representation, which is again estimated through a manifold learning technique that estimates a low-dimensional font space.
More recently, Wang et al.~\cite{wang2021deepvecfont} proposed a dual-modality learning scheme to synthesize vector fonts, which are refined with glyph images using differentiable rasterization.

\mypara{Font stylization}
Efforts have also been made to stylize fonts to enhance their artistic and aesthetic appeal.
For instance, Azadi et al.~\cite{azadi2018multi} proposed a conditional-GAN~\cite{mirza2014conditional,isola2017image} to generate glyph images with different font and texture styles that match a given template.
Berio et al. ~\cite{berio2022strokestyles} proposed to segment a font’s glyphs into a set of overlapping and intersecting strokes to generate artistic stylizations.
To accomplish context-aware text image stylization and synthesis, Yang et al.~\cite{yang2017awesome,yang2018context1,yang2018context2} proposed a style transfer method with the ability to preserve legibility.
Nonetheless, artistic typography surpasses mere manipulation of fonts and often entails imbuing letters with a semantic meaning.

\mypara{Semantic and artistic typography}
Semantic typography involves adding certain elements to a text to emphasize certain aspects, communicate a message, or highlight a property. There have been several endeavors to integrate multiple elements in the creation of a logo, text, or other design elements. One example of this is through the use of collage-based techniques to fill a letter by incorporating semantic elements within it ~\cite{kwan2016pyramid,saputra2019improved,chen2019manufacturable,zhang2022creating}.
The concept of legible calligrams~\cite{zou2016legible} focuses on solving an inverse problem by placing a word or group of letters into a semantic shape. When our approach is applied to an entire word as the input (as shown in Figure~\ref{fig:res_qual_multi_letter}), it may yield results that resemble legible calligrams.
However, our problem statement is entirely distinct, and our approach to solving it is significantly dissimilar because our method is learning-based and not limited to a predetermined template shape that dictates the letter placement and deformation.

The works that are most closely related to ours are those that attempt to alter a letter or a set of letters to achieve a specific semantic meaning, whether through replacement, deformation, or texturization.
For instance, Zhang et al. \cite{zhang2017synthesizing} propose a semi-manual technique that involves manually dividing a letter into sections, fitting semantically related shapes to those sections, and performing post-processing to eliminate artifacts. However, the effectiveness of this method is heavily reliant on the accuracy of manual letter segmentation and the use of predefined shapes, which can impact the quality of the results.
Trick or treat~\cite{tendulkar2019trick} is an attempt to replace a letter with an icon by identifying an icon that closely matches the letter from a joint embedding of letters and icons. The chosen icon is then slightly deformed to better represent the letter. However, this method requires the existence of an icon that closely resembles the letter in order to produce satisfactory results.
Instead, we argue that a more effective solution would involve learning how to \emph{generate} the desired semantic typography by creating letters or words that convey a particular semantic or aesthetic feature in a subtle yet effective manner, as depicted in Figure~\ref{fig:res_qual_single_letter_composed}.

\mypara{Text-based generative design}
Large Language Models such as BERT~\cite{devlin2018bert} significantly advance the understanding of human language, which makes text-based generation tasks much easier.
With the emergence of powerful models that can establish connections between natural language and images, such as CLIP~\cite{radford2021learning}, several downstream tasks have benefited from these models, including mesh and image editing, stylization, and generation~\cite{michel2022text2mesh,jain2022zero,wang2022clip,mildenhall2020nerf}. Of particular relevance to our work, CLIPDraw~\cite{frans2021clipdraw} aims to produce SVG-format drawings by first rasterizing them using DiffVG~\cite{li2020differentiable}, and then utilizing CLIP for evaluation.
Recently, there has been a surge in popularity of diffusion models~\cite{sohl2015diffusion, ho2020ddpm, song2020denoising, song2020improvedsd}, including stable diffusion~\cite{rombach2022high}, which has produced impressive text-to-image results. Our approach utilizes latent diffusion~\cite{rombach2022high} to encode and decode glyph and style images, and employs BERT to condition the denoising process (Figure~\ref{fig:method_pipeline}). To ensure both glyph and style images are respected, we utilize a discriminator to combine adversarial learning and diffusion, making it one of the first of its kind. In contrast to Diffusion-GAN~\cite{wang2022diffusion} which uses a discriminator to distinguish a diffused real image from a diffused fake image at all steps, our discriminator is designed to preserve glyph structures in stylized images as one component of our optimization objectives.

\mypara{Semantic Typography}

In concurrent work, Word-as-Image Semantic Typography (ST) \cite{iluz2023word2img} stylizes a letter through a semantic-aware {\em font deformation\/}; see Figure~\ref{fig:semantic_comp}. 
To guide the deformation, ST uses a pre-trained Stable Diffusion model along with losses to preserve the font structure. 
Our approach differs from ST in multiple ways. We focus on extracting salient features of a style and applying them to a glyph shape and ensure legibility using a discriminator rather than deforming a font. This allows us to incorporate multiple relevant colors to semantic and stylistic attributes in raster form, while the results produced by ST remain single color in vector form. In addition, we can stylize multiple letters together as a single shape (Figure~\ref{fig:res_qual_multi_letter}) while ST deforms each letter individually.

\section{Method}
\label{sec:method}

%\subsection{Overview}

%Embedding semantics and textures into glyphs is not a straightforward task, thus it is not possible to use pixel-level supervision to obtain the results. A more abstract understanding of the semantics is required to successfully transfer them to a glyph. Latent Diffusion Models (LDMs) .... 

%Our key idea is to employ a discriminator on top of LDMs to discriminate between the latent codes of synthesized and vanilla glyphs. The discriminator aims to guide the generator to produce images falling within the glyph shape. Meanwhile, a diffusion loss ensures that the original style and semantics of input style images are not lost. Through this, the generator associates the text prompt with the glyph shape, and the style of input images. The pipeline of our method is demonstrated in Figure~\ref{fig:method_pipeline}. 

%\vspace{7pt}
%
%\mypara{Input and output}
%
Our method takes as input a {\em style prompt\/}, in the form of texts, and a {\em glyph\/} to be stylized, as a raster image. \rz{As we focus on generating artistic typographies in this work, the input glyph represents a graphical form of a letter or a word, e.g., in a particular font.} The output of our method is a stylized version of the glyph based on the style prompt, which consists of a style {\em word\/}, and optionally a style {\em attribute\/}. The style word is a noun specifying an object \rz{or activity} whose semantics are embedded into the resulting typography, while the style attribute provides a further characterization. 
\rz{Compelling typographies can be generated when the input glyph letters are part of the style word, e.g., ``C" in ``cat", 
and the stylized glyph is displayed within the word, resulting in a ``{\em word-as-image\/}'' \cite{berio2022strokestyles,tendulkar2019trick,zhang2017synthesizing,iluz2023word2img}}; see Figure~\ref{fig:teaser}.

We express an input by putting the style prompts in quotations and use the function $\mathscr G(\cdot)$ %\yw{$\mathcal{G(\cdot)}$} 
to denote a mapping from letter contents to glyph images. For example, the running example in Figure~\ref{fig:method_pipeline} has the input, ``cat" + ``cute'' + $\mathscr G$(C). 
%If the whole word is to be stylized, we simply remove the input letters for brevity: ``cat" + ``cute" + $\mathscr G$(CAT).

%If no glyph is provided the method will use the style word as the glyph. 
%Though most of our examples focus on textual input, a glyph can technically be any shape. 
% This glyph is what the style word will be applied to. 

\subsection{Overview}

To generate artistic typography, we first turn the input style word into a visual representation by employing a Latent Diffusion model ~\cite{rombach2022high} to obtain a set of twenty-five style images. The generative task then amounts to embedding the input semantics and the style images into a new, artistically stylized glyph, based on the input font. In the absence of any target images to provide direct supervision, we not only need a suitable feature representation for both the glyph and the style prompts, but also a means to evaluate the stylized results implicitly and effectively.

Utilizing the Latent Diffusion as our architecture backbone, we introduce the key idea of incorporating a {\em discriminator\/}, which guides the Diffusion model to produce images that fully blend styles into the input glyph images. Specifically, the diffusion first constructs the latent space of the given styles and outputs plausible latent codes, then the discriminator aims to distinguish between synthesized results and vanilla glyphs.
Figure~\ref{fig:method_pipeline} shows our method pipeline, with details to follow.

%Firstly, the input style word is used to generate a bunch of style images using a latent diffusion model. The style images can also be collected manually.
%
%Alongside this, we generate a set of rasterizations of the given glyph. Depending on the type of experiment we may use one font or multiple fonts. In either case, the glyphs are rasterized with random colors in each iteration. 

%A more abstract understanding of the semantics and glyphs is required to successfully handle this task.

%The given styles are expected to be naturally blended into the glyph shapes so the discriminator is unable to 
%By this means, the LDM associates the text prompt with the glyph shape, and the style of input images. 
%ensures that the original style and semantics of input style images are not lost
%, for which diffusion models are an excellent choice
%In the meantime, we expect the noises in SD models that produce logo and style images to be as close as possible, to maintain the semantics in logo images.

% In this case, a set of 5-10 should suffice, and their content should match the style word.

%When the inputs are ready, the algorithm works to learn the stylized glyphs. The sample images will be used to help fine-tune the denoising generator. 
\subsection{The Style Latent Space}

To generate a diversity of artistic typography images, we need to construct a latent space of given styles.
Following the LDM, the style images are first fed into an encoder and then passed through a diffusion process. The encoder outputs the features maps $z_{s}$. A Gaussian noise $\epsilon \in \mathcal{N}(0, \mathcal{I})$ is applied on $z_{s}$ to obtain $\bar{z}_{s}$, following an iterative noise \rz{injection} process introduced by diffusion models. 

A {\em denoising generator\/}, with a U-Net~\cite{ronneberger2015u} architecture conditioned on \rz{an encoded vector} by BERT, performs the denoising process for $\bar{z}_{s}$. It aims to predict the added noises $\hat{\epsilon}$ from $\bar{z}_{s}$ and then denoise $\bar{z}_{s}$ to $\hat{z}_{s}$. The prompt serves as a condition to the diffusion process, \rz{to enforce the fusion of the styles and glyphs into our desired images.} 
\rz{Technically, any text-conditioning vector fed to the denoising generator, even a random one, could be applied since over time, the generator will fine-tune it. However, in our current
implementation, we combine the input style prompt with the designated input letters as a prompt to BERT, and use the BERT-encoded vector for text conditioning, with the intent to ``warm start" the generator for faster convergence.}
% For comparison, we show in the ablation study the effect of using a random word as the prompt. }
Finally, we obtain $\bar{z}_{s}$ for sampling styles and $\hat{z}_{s}$ for generating our desired images. During training and inference, $\hat{z}_{s}$ will be sent into the discriminator and the decoder, respectively.

%LDM works on a latent representation of images which makes it far easier on resources and easier to train. 
%Using the predicted noise $\hat{\epsilon}$ and the number of (de)noising steps, LDM outputs the denoised style latent code $\hat{z}_{s}$. 

%Since our task is a sub-task of image generation, it is impossible to train the model from scratch with only a few images. 
The pre-trained encoder and decoder from Latent Diffusion~\cite{rombach2022high} are used in our network; they are frozen during both the training and testing processes. The encoder and decoder were trained with 400M images and can extract high-quality image features.
We employ a diffusion loss which measures the mean square error (MSE) between the predicted noise $\hat{\epsilon}$ and the input noise $\epsilon$: $\mathcal{L}_{diff} = || \hat{\epsilon} - \epsilon ||^2_{2}.$
%\[ \mathcal{L}_{diff} = || \hat{\epsilon} - \epsilon ||^2_{2}. \]

\subsection{The Discriminator}
Since there are no target images available in this task, the generated results can only be supervised implicitly.
It is inappropriate to directly align the generated images with glyph images because the former is highly textured and might have a displacement of glyph outlines. 
To this end, we propose to employ a discriminator in the style of GANs as the examiner. Different from vanilla GANs, the discriminator here takes input as feature maps instead of raw images. The feature maps contain both spatial information and local semantics, which helps the Discriminator to more easily build the correspondence between real and fake inputs. \par
Specifically, the rasterized glyph images are fed into the encoder to obtain the glyph feature maps ${z}_{g}$.
% which are subsequently 
Afterwards, the feature maps of style and glyph images ($\hat{z}_{s}$ and $z_{g}$) are sent into a CNN-based discriminator as fake and real samples, respectively.
The discriminator loss $\mathcal{L}_{dis}$ is Binary Cross-Entropy Loss of real/fake prediction:
\[ \mathcal{L}_{dis} = \log(D({z}_{g})) + \log(1 - D(\hat{z}_{s})), \]
where $D(\cdot)$ denotes the function of the discriminator.
Adversarially training $\mathcal{L}_{dis}$ will fine-tune the denoising generator to adapt the prompt to result incorporating the style of input images but following the shape of glyph. 

\subsection{Overall loss function and result ranking}

The \rz{overall} loss function is composed of the discrimination loss and the diffusion loss.
% \[ \mathcal{L} = \mathcal{L}_{diff} + \lambda * \mathcal{L}_{dis}, \]
We alternately optimize the Discriminator $D$ and the Denoising Generator $G$:

\begin{equation}
\min \limits_{G}\max \limits_{D} (\mathcal{L}_{diff} + \lambda \mathcal{L}_{dis}),
\end{equation}
where $\lambda$ is a hyperparameter that makes a tradeoff between maintaining the shape of letters and incorporating characteristics of style images.
Typically, $\lambda$ works best set less than $1$ and is experimentally set to $0.01$. The effect of different $\lambda$ values on results and training is shown in ablation.
The combination of diffusion loss and discriminator loss results in images that incorporate elements of the style while maintaining a structure similar to the glyph.

\begin{figure}
\centering
  \includegraphics[width=1.0\columnwidth]{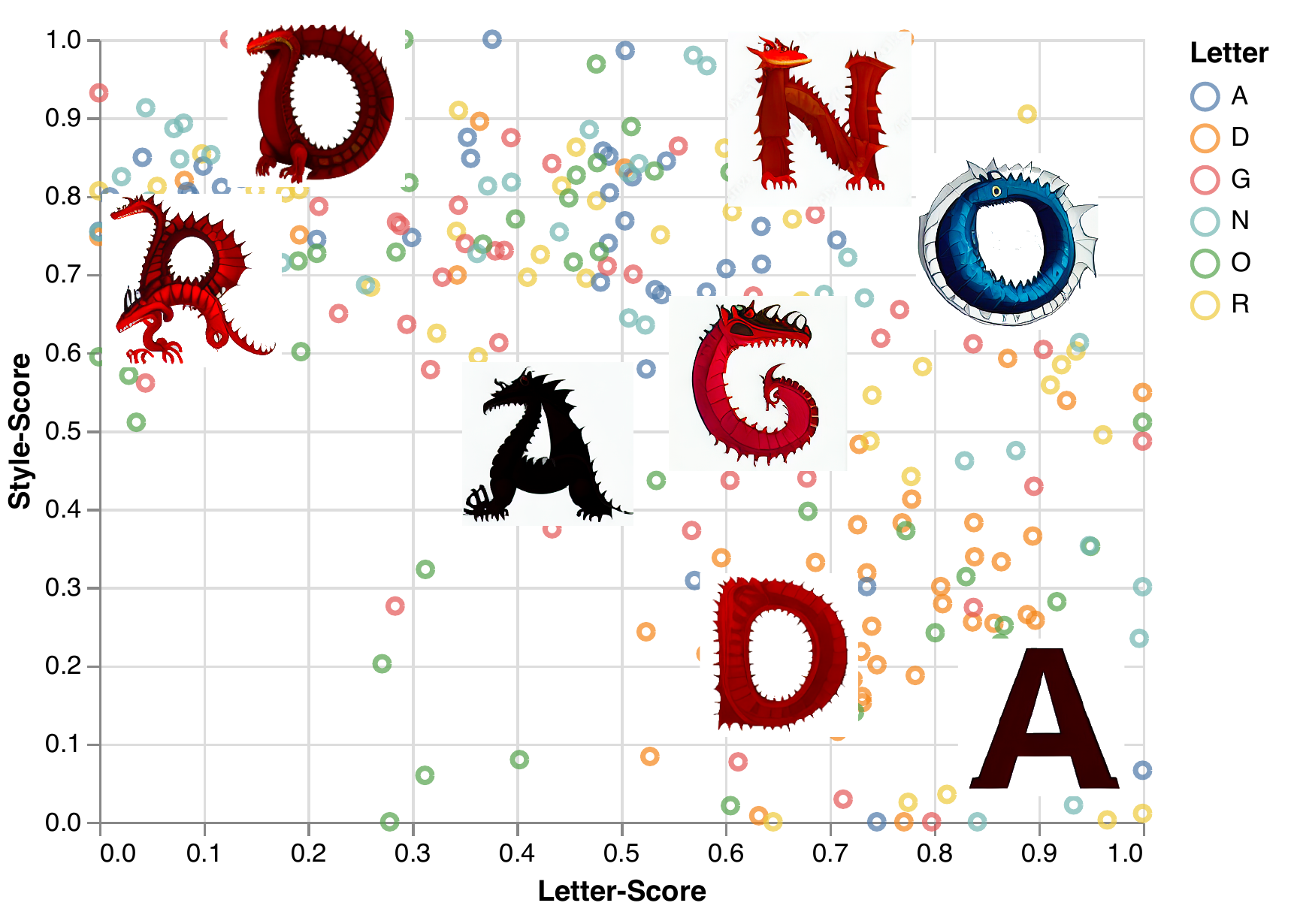}
  \caption{Ranking results. The horizontal and vertical axes respectively denote the scores of glyph and stylistic preservation.}
  \label{fig:ranking}
\end{figure}

%\subsection{Ranking the Generated Results}

Since our model outputs several candidate images from random noises, we have designed a strategy to select better candidates automatically. We employ CLIP to judge the quality of results from both stylistic and glyph preservation standards. Two different prompts are employed to judge each individually: the first is the glyph content like ``Letter R" and the second is style word like ``Dragon".
Figure~\ref{fig:ranking} shows a visual example of ranking a set of stylized glyphs, where the most top-right results are preferred.

\section{Results and evaluation}
\label{sec:results}

To evaluate our DS-Fusion for artistic typography generation, we present qualitative 
results with single-letter and whole-word inputs and show the effects of input fonts, style attributes,
and single- vs.~multi-font training.
%We conduct ablation studies to examine the effectiveness of the discriminator and diffusion losses.
Qualitative and quantitative comparisons are made to the closest alternatives that can be 
adapted to produce similar forms of typographies, including 
DALL-E 2~\cite{ramesh2022hierarchical}, Stable Diffusion (SD)~\cite{rombach2022high}, CLIPDraw~\cite{frans2021clipdraw}, and 
a Google search baseline. As artistic creations are often judged subjectively, we conduct
user studies for our comparative studies including those against searchable contents produced by human artists.

\vspace{7pt}

\mypara{Inputs}
We test our method on input style prompts and glyphs that are applicable to a variety of object categories such as those from animals, 
plants, professions, etc. The input choices reflect an intention to generate compelling artistic typographies, as well as to facilitate comparisons to 
baseline approaches and artist creations.
%In particular
Particularly, where possible, we attempt to produce results using DS-Fusion on inputs picked
in other works to ensure fairness and stress test our method.
On the other hand, the input fonts are randomly selected from Ubuntu built-in font library.

\vspace{7pt}

\mypara{Training details}
For all experiments, we use the pre-trained Latent Diffusion Model with BERT~\cite{rombach2022high},
% We keep the U-NET Encoder/Decoder and BERT frozen, while we fine-tune the denoising generator. 
An Adam optimizer with learning rate of 1e-5 is applied for the denoising generator and for the discriminator,
the learning rate is 1e-4. All the hyper parameters associated with the Latent Model are kept to their default values. 
The complete network, including the discriminator, is fine-tuned to one particular style and glyph combination. 
% For example, to generate "Cat" styled "C", we fine-tune the network on style images of "Cat" and rasterized images of glyph C. After training, the network generates multiple results with the given style and glyph combination.
% The rasterized glyph images are either in the desired font, in the case of one-font mode, or a collection of random fonts in the case of multiple-font mode. In either case, the glyphs are rasterized in random colors.  
We train the network for 800 epochs for one-font mode and 1,200 epochs for multi-font mode. On a NVIDIA GeForce RTX 3090, this takes \(\sim\)5 minutes and \(\sim\)8 minutes, respectively,
to obtain the final results.

% \rz{[RZ: Maham to complete. Emphasize fine-tune/overfit network. Training takes input prompt plus glyph. Experimentally determine
% training epochs: 500 for single-letter and 1,200 for whole-word, etc. used throughout all experiments.]}

\vspace{7pt}

\mypara{Parameters}
There are two free parameters in our method: the number of style images and discriminator weight $\lambda$. Both were set
experimentally and fixed throughout our experiments. We provide an ablation study in Section~\ref{subsec:ablation}.

\vspace{7pt}

\mypara{Evaluation metrics}
%
%The generated typographies are evaluated in terms of their legibility, fidelity, and artistry or creativity. Objectively, we utilize the CLIP loss 
%for \rz{both style and glyph preservation} and MSE Loss for the content word.
Objectively, we evaluate the generated typographies for their legibility via optical character recognition provided by EasyOCR~\cite{jaided2020easyocr} 
and style incorporation via CLIP. Specifically, we use affinity scores calculated by CLIP between the generated typography results and those produced with 
the prompt ``Picture of $<$style word$>$." Subjective tests to additionally assess the artistry and creativity of the results are conducted via user studies.

\subsection{Qualitative results}

\begin{figure*}
\centering
  \includegraphics[width=0.99\textwidth]{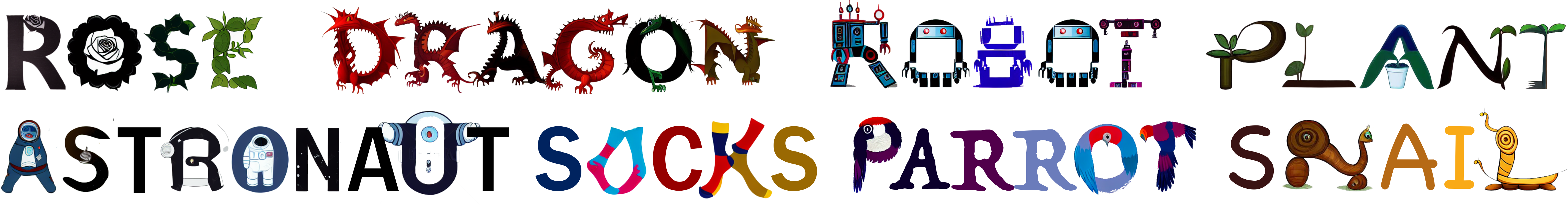}
  \caption{Single-letter artistic typography generated by DS-Fusion, demonstrating the quality and versatility of the stylization based on semantics of the word and the input glyph shapes. First row: all letters are stylized. Second row: selected letters are stylized.}
  \label{fig:res_qual_single_letter_composed}
\end{figure*}

\mypara{Single-letter input}
In Figures~\ref{fig:teaser} and \ref{fig:res_qual_single_letter_composed}, we show a sampler of compelling single-letter artistic typography results generated by our method.
Each result is produced by composing generated single-letter results with renderings of the remaining letters in the input style word. Since the generated results may have 
noisy backgrounds, we employ DeepLab~\cite{chen2017deeplab} for object segmentation to remove them.
To render the other letters, we use the same font as the stylized glyph and select a complementing color which is often the dominant color in the stylized glyph.
\rz{More results can be found in the supplementary material.}

\vspace{7pt}

\begin{figure}
\centering
  \includegraphics[width=0.99\columnwidth]{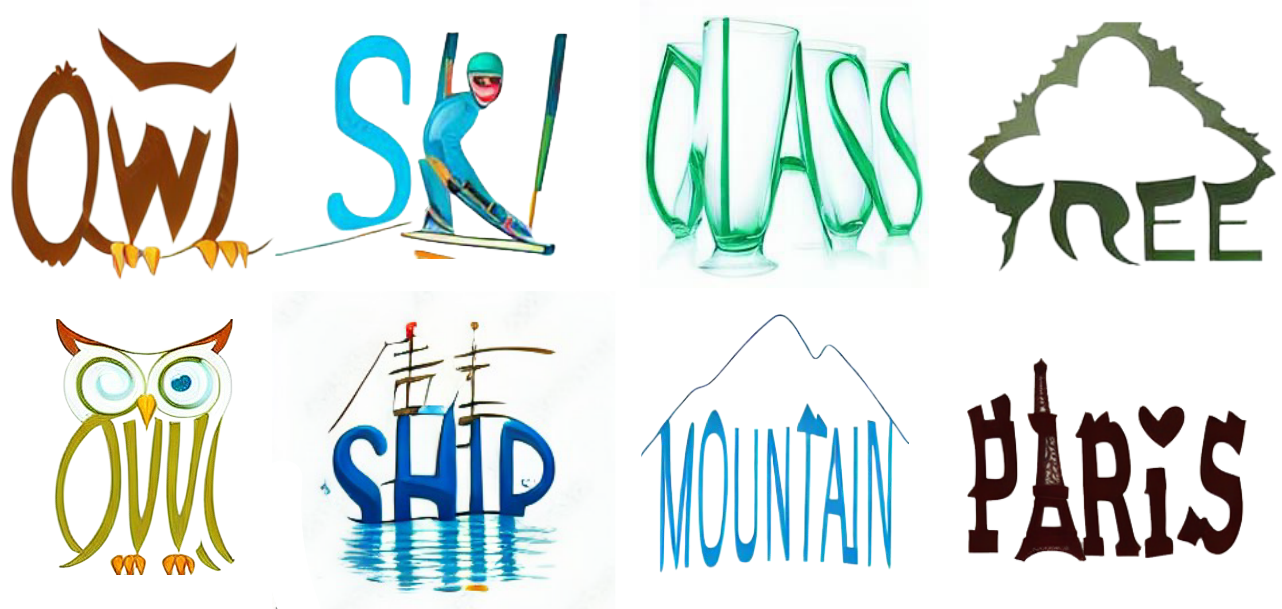}
  \caption{\rz{Results from DS-Fusion to stylize whole words, showing artistic letter combination and auxiliary complementation.}}
  \label{fig:res_qual_multi_letter}
\end{figure}

\mypara{Multi-letter input}
Synthesizing artistic typographies from multiple glyphs is more challenging due to the added structural complexities in the inputs, as well as the more global context to account for. With a higher degree of freedom in the generative process, the tradeoff between legibility and artistry becomes harder to control.
% The method then works by reconstructing style images using diffusion loss while balancing discriminator loss to rasterized images of the content input. 
In Figure~\ref{fig:res_qual_multi_letter}, we show a sampler of results from {\em whole-word\/} stylization.

In more ways than one, our DS-Fusion demonstrates its ability to utilize all the letters of a word to convey semantic features, in a creative manner. 
Particularly compelling are the two results for the style word ``OWL''; the owls' bodies are well formed by the stylized glyphs. 
Auxiliary stylizations can also be generated to offer a global context, e.g., the tree and mountain tops, while the ship image shows the word ``SHIP"
replacing the vessel itself, even including a reflection in water. On the other hand, stylization could also be applied to individual letters, e.g., for ``PARIS."

%\subsection{Effect of Input Fonts and Style Images}

\vspace{7pt}

\begin{figure}
\centering
  \includegraphics[width=0.99\columnwidth]{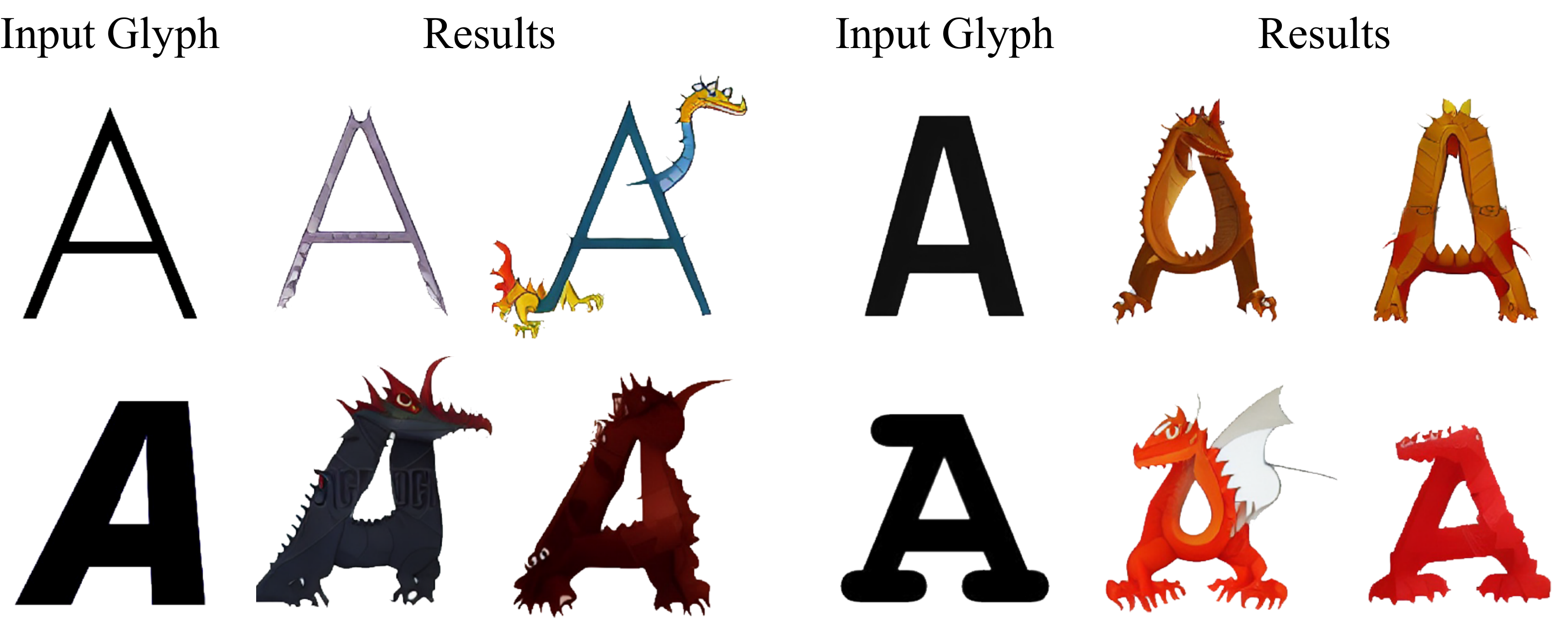}
  \caption{\rz{Details in the input fonts, e.g., thickness, slanting, and accents, are well reflected in results produced by our method.}}
  \label{fig:effect_of_fonts}
\end{figure}

\mypara{Effect of input fonts}
Figure~\ref{fig:effect_of_fonts} shows results produced by varying the fonts of the input glyphs, for ``dragon" + $\mathscr G$(A). They highlight the ability of our method
to preserve various shape characteristics of the input, such as stroke thickness, slanting, and even small-scale details such as the accent on top of the letter stylized as dragon heads.

\vspace{7pt}

\begin{figure}
\centering
  \includegraphics[width=0.99\columnwidth]{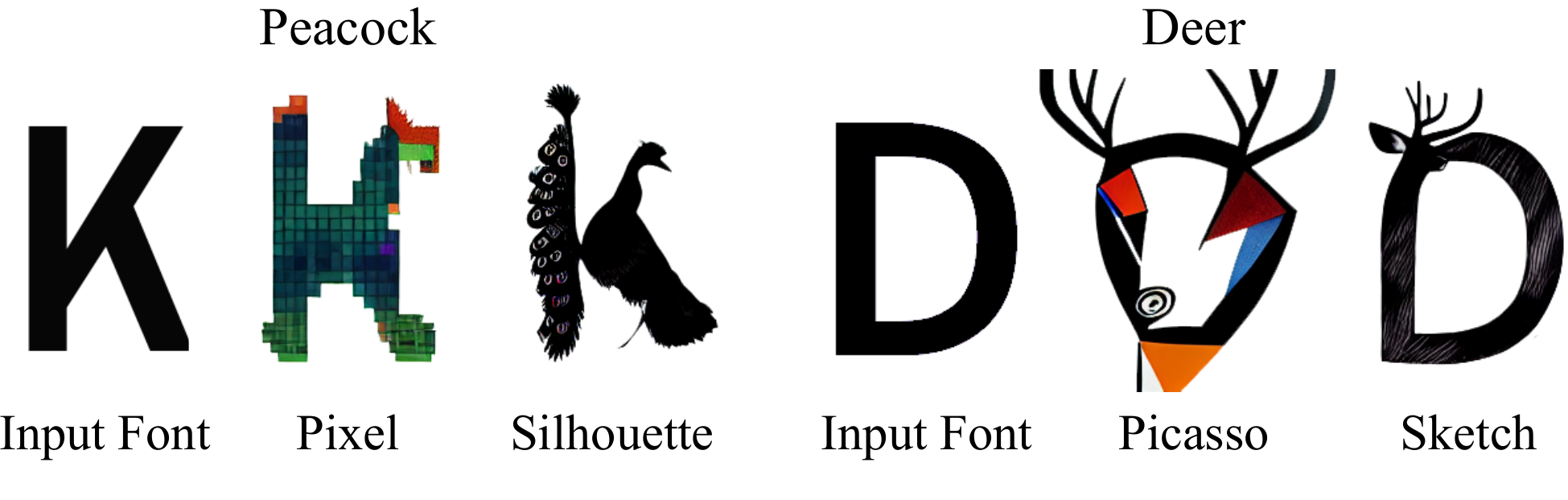}
  \caption{Additional style attributes such as ``pixel", ``sketch", and ``Picasso" can be well reflected in our results.}
  \label{fig:effect_of_styles}
\end{figure}

\mypara{Effect of style attributes}
%
%The style images are generated from Latent Diffusion model. Our results are generated using prompt ``$<$style\_word$>$ cartoon" or ``$<$style\_word$>$'' illustration". This is because images with these properties tend to have a good transferable style to text and logos.  However other prompts like further adjectives like pixel, realistic, etc can also be used. We show the effect of these prompts 
The use of style attributes in the input prompts can be an effective means to further fine-tune the stylization, as shown in Figure~\ref{fig:effect_of_styles}. 

\vspace{7pt}

\mypara{Single- vs.~multiple-font training}
In single-font mode, we pick one random font out of a set of five selected fonts. During training, we change the color of the font randomly. Results of the single-font training show a high degree of  shape correlation to the input font, as shown in Figure~\ref{fig:effect_of_fonts}.

In multiple-font mode, we pick both a random font and a random color in each training step. It takes longer for the model to converge, but it can produce more abstract results, as reflected by the comparisons in Figure~\ref{fig:effect_of_multiple}. Note that we cannot control the font shape in this mode, however, it follows the general shape of the input glyph.

\subsection{Comparisons}

% \input{tables/tab_ablation.tex}

% Comparison using CLIP vs BERT to compute conditions 
\begin{figure*}[!htb]
\centering
  \includegraphics[width=0.97\textwidth]{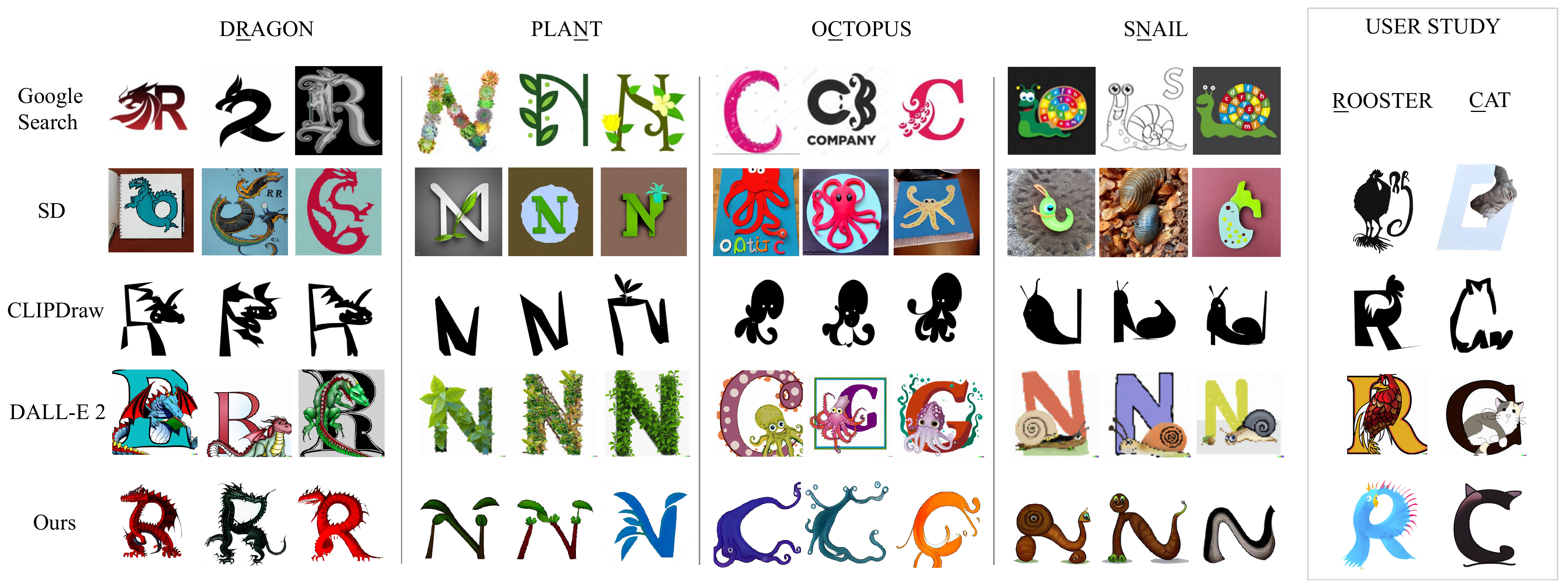}
  \caption{Visual comparison on single-letter (underlined) generation between different methods, with three most compelling results picked per method. The last two columns show sample results from the first user study (Section~\ref{subsec:user}), where Google search was not considered.}
  \label{fig:res_visual_comp}
\end{figure*}

\mypara{Qualitative comparisons}
We compare DS-Fusion to DALL-E 2~\cite{ramesh2022hierarchical}, vanilla Stable Diffusion (SD)~\cite{rombach2022high}, Google Search, and CLIPDraw~\cite{frans2021clipdraw} for which we modify its input from random strokes to vector outlines of the input glyphs to fit the task at hand.
As we can see from the visual comparisons in Figure~\ref{fig:res_visual_comp}, in most cases, DALL-E 2 cannot blend the semantics into the letters; the objects simply surround the letters. Results from SD usually do not contain the shape of the input letters. Google search could occasionally return good results only because the input prompts have had corresponding artist designs that were retrieved. For less common queries such as ``Snail in letter N," retrieval clearly cannot work. As for CLIPDraw, neither the content legibility nor the semantic embedding is satisfactory.

\begin{figure}
\centering
  \includegraphics[width=0.99\columnwidth]{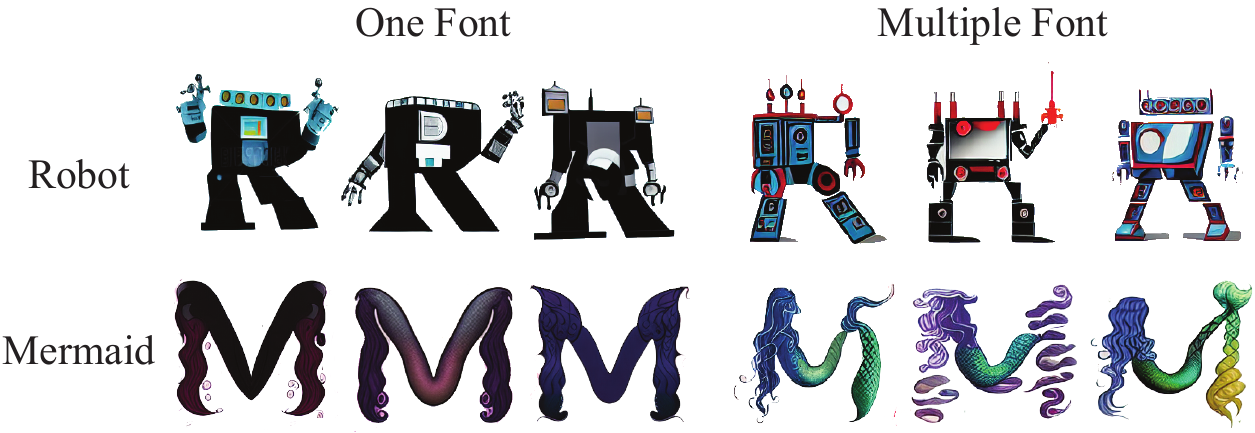}
  \caption{One-font vs.~multi-font training. Results show a stronger legibility in the former, and a more abstract stylization in the latter.}
  \label{fig:effect_of_multiple}
\end{figure}

\vspace{7pt}

\mypara{Quantitative comparisons}
To evaluate the performance of different methods in terms of legibility and style incorporation, we tested on 21 style words across a wide range of semantic categories (see supplementary material for the list) and stylized all the letters from them. Since both CLIPDraw and our method can receive multiple fonts as input, we also separately report their performances when receiving single and multiple fonts. For each style word, 4 {\em random\/} results were sampled from each method to obtain statistics.

\begin{table}
	\centering
	\begin{tabular}{lccc}
	\toprule
        Method & OCR$\uparrow$ & OCR [blurred]$\uparrow$  & CLIP$\uparrow$ \\
		\midrule
             DALL-E 2~\cite{ramesh2022hierarchical} &  4.9 & 9.8 &  22.3\\
             SD~\cite{rombach2022high}  &  7.0 & 4.9 &  \textbf{23.1}\\
             CLIPDraw~\cite{frans2021clipdraw} (SF) & 4.6 &  15.8  &  22.7 \\
             % Google Search &  &  &  \\
             DS-Fusion (SF)  & \textbf{36.4} &  \textbf{61.8} & 21.1\\
             \midrule
             CLIPDraw~\cite{frans2021clipdraw} (MF) &\textbf{6.4}&  17.9 &  23.2\\
             %DF-GAN~\cite{wang2022diffusion} &  &  &    \\
             DS-Fusion (MF)  &  5.1 & \textbf{20.5} & \textbf{24.1}\\

		\bottomrule
	\end{tabular}

  	\caption{Quantitative comparisons between different methods in terms of legibility (OCR) and style incorporation (CLIP). ``SF" and ``MF" denote single- and multi-font inputs, respectively.}

    \label{tab:eval_one_font}
\end{table}

The quantitative results are shown in Table~\ref{tab:eval_one_font}.
On single-font inputs, our method significantly outperforms the others in OCR accuracy both for raw and blurred versions of the generated results. The style score of DS-Fusion is slightly behind the others, which is not unexpected since in one-font mode, the output favors the glyph, and the style is less pronounced; see Figure~\ref{fig:effect_of_multiple}. Since CLIPDraw is over-fitted to the CLIP loss, its style score tends to be high while the corresponding visual results are not satisfactory; see Figure~\ref{fig:res_visual_comp} and the following user study.
When taking multiple fonts as input, DS-Fusion achieves the best score ($24.1$) of style incorporation among all methods. The letters are stylized in a much more abstract fashion, which also makes them more difficult to be recognized. Nevertheless, the OCR accuracy of DS-Fusion is comparable to CLIPDraw. 

% with one-font and multiple fonts with eighteen and fourteen categories  respectively
% Also, CLIP shows our style is also within the range of top methods, especially as CLIPDraw has been finetuned on CLIP. 
%Our results show that not only are the typographies legible, but the style score is also close to other methods like DALL-E-2, CLIPDraw and Stable Diffusion.
%multiple fonts as input on another 14 categories.
%For our results, we also employ our ranking method to automatically pick the top four out of fifty generated samples for evaluation.\mt{[we are not doing this anymore. our samples are also random]}

% \vspace{7pt}

% \mypara{Timing}
%
%\ali{is this supposed to be removed?}

\vspace{7pt}

\begin{figure}
\centering
  \includegraphics[width=0.99\columnwidth]{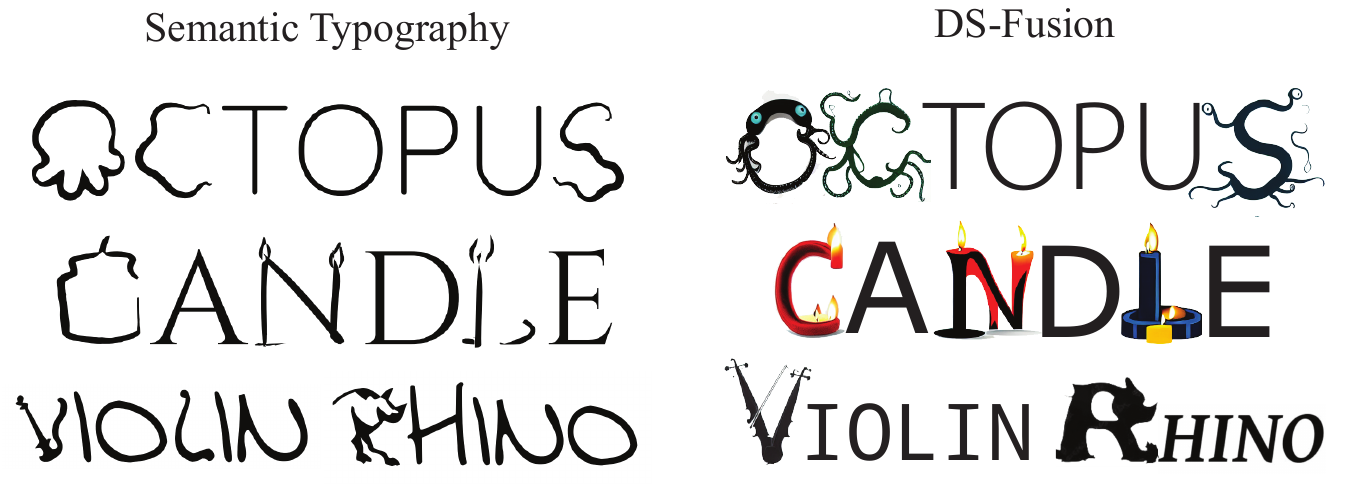}
  \caption{Visual comparison with semantic typography~\cite{iluz2023word2img}.}
  \label{fig:semantic_comp}
\end{figure}

\mypara{Comparison with Semantic Typography}
We compare DS-Fusion to Word-as-Image Semantic Typography~\cite{iluz2023word2img} in Figure~\ref{fig:semantic_comp}. both methods perform similarly; both results are still readable and semantically relevant (e.g., rhino, violin). However, our method is capable of adding colorful and artistic textures to make the results more semantically relevant and visually appealing, e.g., candle and octopus.

\subsection{User Study}
\label{subsec:user}

A user study serves to evaluate our method via subjective human judgement.
%The study has two parts
We perform two such studies: in the first one, users select between DS-Fusion and other generative methods; in the second, a different group of participants selects between DS-Fusion and artist designs.
%We compare with both results from other generative methods and human artists.
Both studies start by showing participants the definition and examples of artistic typography, which give them an idea of what to expect.
We directly chose {\em artist-created\/} examples from a popular online tutorial on \href{https://amadine.com/useful-articles/create-artistic-typography-designs-with-amadine}{artistic typography design},
 %recommended by Google, 
instead of picking inputs that may favour our method. \rz{The same 10} inputs with distinguishable style words (e.g.,\underline{C}at, \underline{P}arrot, etc.) were used in both studies.

The first study compares our results to those from CLIPDraw, DALL-E 2, and SD.
To prepare the study, we asked a professional designer to select the most representative result for each of the four candidate results generated by each method. The study collected responses from 32 participants with different occupations and varying artistic backgrounds to determine which result was deemed best. Table~\ref{tab:eval_user_study_vs_others} shows that DS-Fusion significantly outperformed other methods. 

While DALL-E 2 also gathered close to 30\% votes, some of its results preferred by users do not contain any stylization of the input letters, e.g., see the last two columns of Figure~\ref{fig:res_visual_comp}. DALL-E 2 simply placed an image of a cat or rooster near an un-stylized letter. Clearly, these are not satisfactory typography results, yet user subjectivity has led to them receiving 25\% and {\bf 50\%} of the votes, respectively.

%In the first \am{study}, we compare with synthetic results from CLIPDraw, DALL-E 2, and Stable Diffusion. \yw{For each question, we  generated \am{four} candidate results from each method. Then, we invited a professional designer to %determine which one is 
%\am{select}
%the most representative result for each method. Afterwards, we passed the user study to
%wide range of 
%\am{31} users from different occupations
%\am{to select the method producing the best result.}
%The users will select which method produces the best result in each question. 
%There are 31 participants in this part.
%\am{Table.~\ref{tab:eval_user_study_vs_others} shows the users' preference, where we outperform all the methods by a large margin.}}
%shows the percentage of user preference for each method, where our results are the most favorite, leading the second DALL-E 2 by a large margin.}\par

The second 
%part of our 
study collected responses from 42 participants to choose between our results and professionally crafted examples found in the tutorial. %the users select a better one from two options based on their understanding of artistic typography. 
%There are 40 participants in this part. 
The study results are shown in the second part in Table~\ref{tab:eval_user_study_vs_others}. %Considering that human-designed results rely heavily on the creativity, expertise and efforts, it is impressive that 
In about 42\% of the cases, users found our results better or equivalent to that human-designed examples. This is a satisfactory outcome since human-designed examples heavily rely on the creativity and expertise of designers (more details about our user study in the supplementary material).

\begin{table}
	\centering
	\begin{tabular}{ccccc}
	\toprule
        CLIPDraw & DALL-E 2 & SD  & Equal & DS-Fusion \\
		\midrule
            6.88 & 29.38 & 9.38  & 5.31 & \textbf{49.06} \\\hline\hline

            \multicolumn{2}{c}{Human} & \multicolumn{1}{c}{Equal} & \multicolumn{2}{c}{DS-Fusion}\\
            \midrule
            \multicolumn{2}{c}{57.14} & \multicolumn{1}{c}{11.19} &  \multicolumn{2}{c}{31.67} \\
		\bottomrule
	\end{tabular}

  	\caption{User Study comparing DS-Fusion with other generative methods (Top) and human designs (Bottom). The numbers show the percentage of user preference for different results. ``SD'' denotes Stable Diffusion. ``Equal'' denotes equally good.}

    \label{tab:eval_user_study_vs_others}
\end{table}

\subsection{Ablation studies}
\label{subsec:ablation}

\begin{figure}
\centering
  \includegraphics[width=0.97\columnwidth]{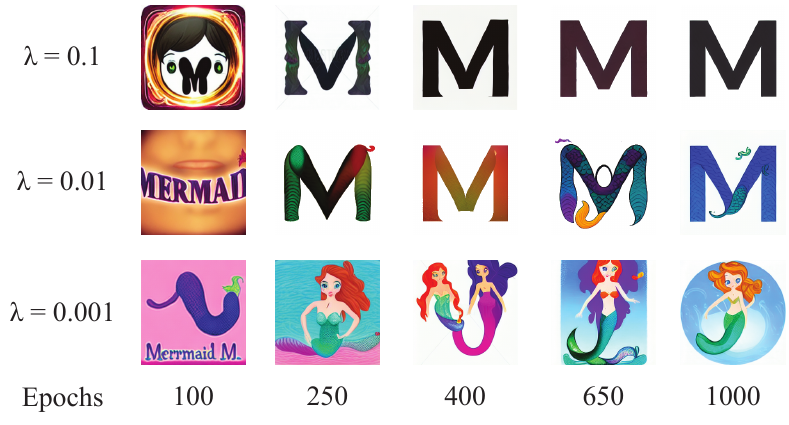}
  \caption{Effect of discriminator weight, on ``Mermaid" + $\mathscr G$(M).}
  \label{fig:ablation_lambda}
\end{figure}

\mypara{Discriminator weight}
%
%In Figure~\ref{fig:ablation_lambda} we show how changing the  weight of Discriminator loss i.e. the  $\lambda$ value affects the training process.  
The impact of the $\lambda$ value on adjusting the Discriminator loss is shown in Figure~\ref{fig:ablation_lambda}.
% We also show the training results if we remove the diffusion loss altogether. 
A high value of $\lambda$ pushes the generator towards the glyph shape, 
%and vice versa 
%A small value results in the generator either taking a long time to converge to the glyph shape or it is not able to reach it at all.
However, the generator may take a long time to converge to the glyph shape or may not be able to reach it at all when a small value of lambda is used.
Therefore, the value of lambda was selected experimentally. 
%On the other hand, 
%if we eliminate the diffusion loss,
% \am{AlsTro, if the diffusion loss is removed,}
% the generator quickly forgets the semantics of the style images corresponding to our prompt.

% \vspace{7pt}

% \mypara{Conditioning vector}
% %
% In Figure ~\ref{fig:alblation_prompt} we show the effect of using alternative prompts as conditioning for the denoising generator. We show by using a random prompt, which is a five-length random character string, how the method converges over time. However, using a prompt that is a better representative of our style images greatly speeds up and refines the process.

% \input{figs/fig_ablation_prompt}

\vspace{7pt}

\begin{figure}
\centering
  \includegraphics[width=0.97\columnwidth]{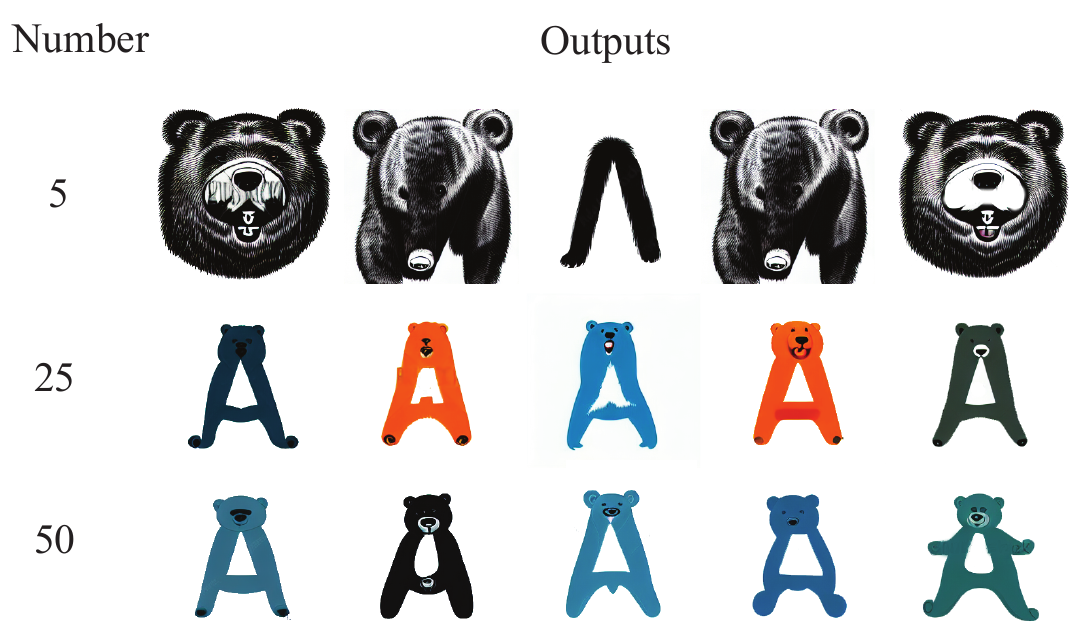}
  \caption{Effect of generating a different number of style images. The left column shows the number of images generated, and on the right, we see the range of output results. }
  \label{fig:ablation_styimg}
\end{figure}

\mypara{Number of style images}
In Figure ~\ref{fig:ablation_styimg} we show the effect of generating different numbers of style images. When the number is too low, the method
%finds it difficult to 
struggles to
extract common features to apply to the glyph. As the number of images increases, we see not only improvement in generation quality but also  in the diversity of outputs. 
We do not observe a significant improvement in increasing the value excessively, hence we opt for the number 25 as an optimal balance.
%However, we don't see a significant advantage going too high, and therefore choose the number twenty-five as a good balance.

\vspace{7pt}

\begin{figure}
\centering
  \includegraphics[width=0.97\columnwidth]{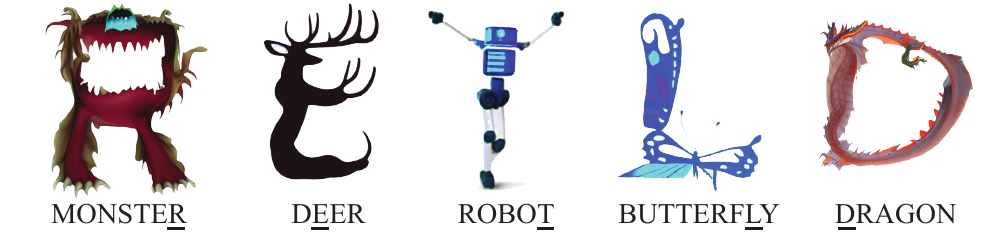}
  \caption{Results on using CLIP encoder for text conditioning.}
  \label{fig:clip_results}
\end{figure}

\mypara{Text encoder}
Our results are generated using BERT. This model has a smaller resolution and is faster to generate compared to CLIP SD.
%makes it faster to develop and generate results. 
However, we show some results using the CLIP text encoder in Figure~\ref{fig:clip_results} to demonstrate the generality of our method.
%verify that our method is applicable to other text encoders.

% 
% The quantitative results are shown in Table~\ref{tab:eval_quan_abc}.
% \input{tables/tab_eval_quan.tex}

\section{Conclusion, limitation, and future work}
\label{sec:future}

%\am{We present a method to automatically generate artistic typography for a given letter or a word. This is an extremely challenging task since one cannot simply deform a given letter to a target semantic in an artistic way as prior knowledge about the visual domain of the semantic is necessary. To address this challenge, we incorporated language-based generative models in our approach. Through several quantitative and qualitative experiments, along with user studies, we demonstrated the efficacy of our method and showed that solely using state-of-the-art language-based generative models will not produce satisfactory results.
%Our method is one of the first attempts to combine adversarial learning and diffusion in a single framework. 
%Using diffusion's powerful capabilities for generation and having a discriminator, we can ensure that our generated artistic typography stays faithful to the given letter.}

We developed an automatic method to generate artistic typography for a letter or word by using language-based generative models equipped with a discriminator. We demonstrated the effectiveness of our method, coined DS-Fusion, through extensive experiments and user studies. Our approach combines adversarial learning and diffusion, which helps ensure the fidelity of the generated typography.

%which is a difficult task as one cannot simply deform a given letter to a target semantic in an artistic way as prior knowledge about the visual domain of the semantic is necessary.

% We have presented a method to generate artistic typography for a given letter or a word. This is an extremely challenging task since one cannot simply deform a given letter to a target semantic in an artistic way as prior knowledge about the visual domain of the semantic is necessary. To address this challenge, we incorporated language-based generative models in our approach. Through several quantitative and qualitative experiments, along with user studies, we demonstrated the efficacy of our method and showed that solely using state-of-the-art language-based generative models will not produce satisfactory results.
% Our method is one of the first attempts to combine adversarial learning and diffusion in a single framework. 
% Using diffusion's powerful capabilities for generation and having a discriminator, we can ensure that our generated artistic typography stays faithful to the given letter.

\begin{figure}
\centering
  \includegraphics[width=0.99\columnwidth]{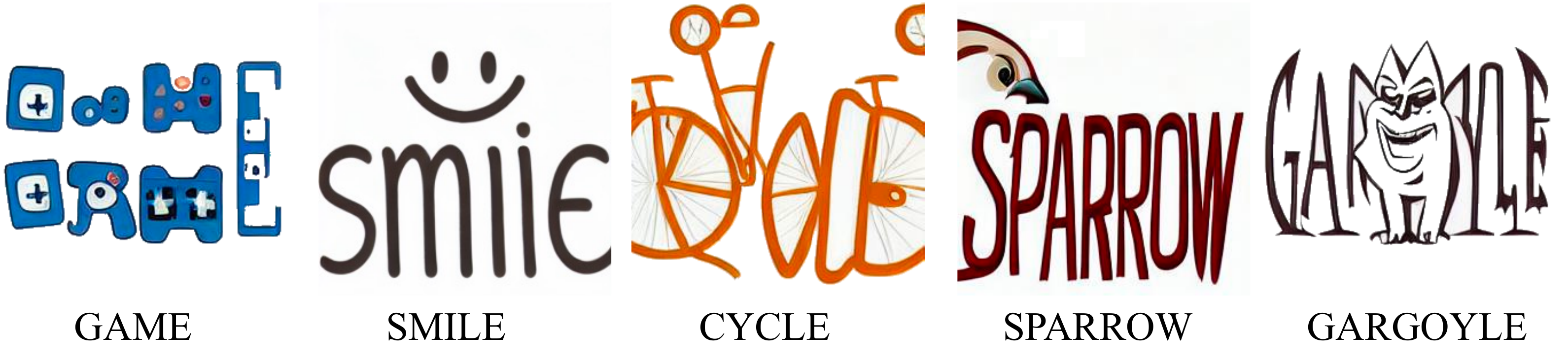}
  \caption{Some failure cases on whole-word inputs.}
  \label{fig:res_limitations}
\end{figure}

Our method is generally effective but it still has some limitations that require further investigation. When dealing with multi-letter inputs, our method may struggle to generate satisfactory results if the style images and letters are too dissimilar (Figure~\ref{fig:res_limitations}).
At present, our method is optimized for each specific combination of style and glyph. However, future work could involve training a network for a particular style that can generate any letter during inference.
Despite having an automatic selection strategy, it may not always generate the most visually plausible outcome. Rather, it can generate a range of plausible results to choose from. A stronger selection mechanism could be a future work.

Our glyph is potentially an image in various forms, including alphanumeric fonts, foreign language characters, or even a 2D shape. It is interesting to creatively modify such shapes to display a semantic (e.g., a flower-shaped chair). 
In addition, for personalization, style images can be manually prepared or drawn if desired. We explored these ideas and presented preliminary results in the supplementary material, but further research is necessary to solidify the outcomes.

% \vspace{7pt}

% \mypara{Glyph as general shape}
% %
% \rz{In general, a glyph depicts a graphic form that can be an alphabetic or numeric font, a character or script from a foreign language such as Chinese, or a combination of them. In its most general form, our input glyph can represent a general 2D shape.}
% As shown in Figure~\ref{fig:res_extensions}, we can transfer the given styles onto other objects like chair, toys and cars etc. 
% \input{figs/fig_extension.tex}

% \vspace{7pt}

% \mypara{Personalized stylization}
% %
% If desired, these style images could also be prepared manually, e.g., for personalization.

%Furthermore, because the method fine-tunes the generator to output results in a specific shape, it can affect prompts other than those used in the method. Therefore, the method should only be used to generate results as designed, otherwise, unexpected outputs might be generated. This can be extended to future work, where the whole generator can be adapted to a specific shape or glyph. However, we did not explore this extension currently.

\clearpage
{\small
\bibliographystyle{ieee_fullname}
\bibliography{egbib}

\begin{thebibliography}{10}\itemsep=-1pt

\bibitem{azadi2018multi}
Samaneh Azadi, Matthew Fisher, Vladimir~G Kim, Zhaowen Wang, Eli Shechtman, and
  Trevor Darrell.
\newblock Multi-content gan for few-shot font style transfer.
\newblock In {\em Proceedings of the IEEE conference on computer vision and
  pattern recognition}, pages 7564--7573, 2018.

\bibitem{balashova2019learning}
Elena Balashova, Amit~H Bermano, Vladimir~G Kim, Stephen DiVerdi, Aaron
  Hertzmann, and Thomas Funkhouser.
\newblock Learning a stroke-based representation for fonts.
\newblock In {\em Computer Graphics Forum}, volume~38, pages 429--442. Wiley
  Online Library, 2019.

\bibitem{berio2022strokestyles}
Daniel Berio, Frederic~Fol Leymarie, Paul Asente, and Jose Echevarria.
\newblock Strokestyles: Stroke-based segmentation and stylization of fonts.
\newblock {\em ACM Transactions on Graphics (TOG)}, 41(3):1--21, 2022.

\bibitem{campbell2014learning}
Neill~DF Campbell and Jan Kautz.
\newblock Learning a manifold of fonts.
\newblock {\em ACM Transactions on Graphics (TOG)}, 33(4):1--11, 2014.

\bibitem{chen2017deeplab}
Liang-Chieh Chen, George Papandreou, Iasonas Kokkinos, Kevin Murphy, and Alan~L
  Yuille.
\newblock Deeplab: Semantic image segmentation with deep convolutional nets,
  atrous convolution, and fully connected crfs.
\newblock {\em IEEE transactions on pattern analysis and machine intelligence},
  40(4):834--848, 2017.

\bibitem{chen2019manufacturable}
Minghai Chen, Fan Xu, and Lin Lu.
\newblock Manufacturable pattern collage along a boundary.
\newblock {\em Computational Visual Media}, 5:293--302, 2019.

\bibitem{devlin2018bert}
Jacob Devlin, Ming-Wei Chang, Kenton Lee, and Kristina Toutanova.
\newblock Bert: Pre-training of deep bidirectional transformers for language
  understanding.
\newblock {\em Proceedings of the 2019 Conference of the North American Chapter
  of the Association for Computational Linguistics: Human Language
  Technologies, Volume 1 (Long and Short Papers)}, pages 4171--4186, 2019.

\bibitem{frans2021clipdraw}
Kevin Frans, Lisa~B Soros, and Olaf Witkowski.
\newblock Clipdraw: Exploring text-to-drawing synthesis through language-image
  encoders.
\newblock {\em arXiv preprint arXiv:2106.14843}.

\bibitem{ho2020ddpm}
Jonathan Ho, Ajay Jain, and Pieter Abbeel.
\newblock Denoising diffusion probabilistic models.
\newblock In {\em NeurIPS}, 2020.

\bibitem{iluz2023word2img}
Shir Iluz, Yael Vinker, Amir Hertz, Daniel Berio, Daniel Cohen-Or, and Ariel
  Shamir.
\newblock Word-as-image for semantic typography.
\newblock {\em arXiv preprint arXiv:2303.01818}, 2023.

\bibitem{isola2017image}
Phillip Isola, Jun-Yan Zhu, Tinghui Zhou, and Alexei~A Efros.
\newblock Image-to-image translation with conditional adversarial networks.
\newblock In {\em Proceedings of the IEEE conference on computer vision and
  pattern recognition}, pages 1125--1134, 2017.

\bibitem{jaided2020easyocr}
AI Jaided.
\newblock Easyocr.
\newblock {\em Retrieved October}, 9:2020, 2020.

\bibitem{jain2022zero}
Ajay Jain, Ben Mildenhall, Jonathan~T Barron, Pieter Abbeel, and Ben Poole.
\newblock Zero-shot text-guided object generation with dream fields.
\newblock In {\em Proceedings of the IEEE/CVF Conference on Computer Vision and
  Pattern Recognition}, pages 867--876, 2022.

\bibitem{kwan2016pyramid}
Kin~Chung Kwan, Lok~Tsun Sinn, Chu Han, Tien-Tsin Wong, and Chi-Wing Fu.
\newblock Pyramid of arclength descriptor for generating collage of shapes.
\newblock {\em ACM Trans. Graph.}, 35(6):229--1, 2016.

\bibitem{li2020differentiable}
Tzu-Mao Li, Michal Luk{\'a}{\v{c}}, Micha{\"e}l Gharbi, and Jonathan
  Ragan-Kelley.
\newblock Differentiable vector graphics rasterization for editing and
  learning.
\newblock {\em ACM Transactions on Graphics (TOG)}, 39(6):1--15, 2020.

\bibitem{michel2022text2mesh}
Oscar Michel, Roi Bar-On, Richard Liu, Sagie Benaim, and Rana Hanocka.
\newblock Text2mesh: Text-driven neural stylization for meshes.
\newblock In {\em Proceedings of the IEEE/CVF Conference on Computer Vision and
  Pattern Recognition}, pages 13492--13502, 2022.

\bibitem{mildenhall2020nerf}
B Mildenhall, PP Srinivasan, M Tancik, JT Barron, R Ramamoorthi, and R Ng.
\newblock Nerf: Representing scenes as neural radiance fields for view
  synthesis.
\newblock In {\em European conference on computer vision}, 2020.

\bibitem{mirza2014conditional}
Mehdi Mirza and Simon Osindero.
\newblock Conditional generative adversarial nets.
\newblock {\em arXiv preprint arXiv:1411.1784}, 2014.

\bibitem{radford2021learning}
Alec Radford, Jong~Wook Kim, Chris Hallacy, Aditya Ramesh, Gabriel Goh,
  Sandhini Agarwal, Girish Sastry, Amanda Askell, Pamela Mishkin, Jack Clark,
  et~al.
\newblock Learning transferable visual models from natural language
  supervision.
\newblock In {\em International conference on machine learning}, pages
  8748--8763. PMLR, 2021.

\bibitem{ramesh2022hierarchical}
Aditya Ramesh, Prafulla Dhariwal, Alex Nichol, Casey Chu, and Mark Chen.
\newblock Hierarchical text-conditional image generation with clip latents.
\newblock {\em arXiv preprint arXiv:2204.06125}, 2022.

\bibitem{rombach2022high}
Robin Rombach, Andreas Blattmann, Dominik Lorenz, Patrick Esser, and Bj{\"o}rn
  Ommer.
\newblock High-resolution image synthesis with latent diffusion models.
\newblock In {\em Proceedings of the IEEE/CVF Conference on Computer Vision and
  Pattern Recognition}, pages 10684--10695, 2022.

\bibitem{ronneberger2015u}
Olaf Ronneberger, Philipp Fischer, and Thomas Brox.
\newblock U-net: Convolutional networks for biomedical image segmentation.
\newblock In {\em Medical Image Computing and Computer-Assisted
  Intervention--MICCAI 2015: 18th International Conference, Munich, Germany,
  October 5-9, 2015, Proceedings, Part III 18}, pages 234--241. Springer, 2015.

\bibitem{saputra2019improved}
Reza~Adhitya Saputra, Craig~S Kaplan, and Paul Asente.
\newblock Improved deformation-driven element packing with repulsionpak.
\newblock {\em IEEE transactions on visualization and computer graphics},
  27(4):2396--2408, 2019.

\bibitem{sohl2015diffusion}
Jascha Sohl-Dickstein, Eric Weiss, Niru Maheswaranathan, and Surya Ganguli.
\newblock Deep unsupervised learning using nonequilibrium thermodynamics.
\newblock In Francis Bach and David Blei, editors, {\em Proceedings of the 32nd
  International Conference on Machine Learning}, volume~37 of {\em Proceedings
  of Machine Learning Research}, pages 2256--2265, Lille, France, 07--09 Jul
  2015. PMLR.

\bibitem{song2020denoising}
Jiaming Song, Chenlin Meng, and Stefano Ermon.
\newblock Denoising diffusion implicit models.
\newblock {\em arXiv:2010.02502}, October 2020.

\bibitem{song2020improvedsd}
Yang Song and Stefano Ermon.
\newblock Improved techniques for training score-based generative models.
\newblock In {\em Proceedings of the 34th International Conference on Neural
  Information Processing Systems}, NIPS'20, Red Hook, NY, USA, 2020. Curran
  Associates Inc.

\bibitem{tendulkar2019trick}
Purva Tendulkar, Kalpesh Krishna, Ramprasaath~R Selvaraju, and Devi Parikh.
\newblock Trick or treat: Thematic reinforcement for artistic typography.
\newblock {\em arXiv preprint arXiv:1903.07820}, 2019.

\bibitem{wang2022clip}
Can Wang, Menglei Chai, Mingming He, Dongdong Chen, and Jing Liao.
\newblock Clip-nerf: Text-and-image driven manipulation of neural radiance
  fields.
\newblock In {\em Proceedings of the IEEE/CVF Conference on Computer Vision and
  Pattern Recognition}, pages 3835--3844, 2022.

\bibitem{wang2021deepvecfont}
Yizhi Wang and Zhouhui Lian.
\newblock Deepvecfont: synthesizing high-quality vector fonts via dual-modality
  learning.
\newblock {\em ACM Transactions on Graphics (TOG)}, 40(6):1--15, 2021.

\bibitem{wang2022diffusion}
Zhendong Wang, Huangjie Zheng, Pengcheng He, Weizhu Chen, and Mingyuan Zhou.
\newblock Diffusion-gan: Training gans with diffusion.
\newblock {\em arXiv preprint arXiv:2206.02262}, 2022.

\bibitem{yang2017awesome}
Shuai Yang, Jiaying Liu, Zhouhui Lian, and Zongming Guo.
\newblock Awesome typography: Statistics-based text effects transfer.
\newblock In {\em Proceedings of the IEEE Conference on Computer Vision and
  Pattern Recognition}, pages 7464--7473, 2017.

\bibitem{yang2018context2}
Shuai Yang, Jiaying Liu, Wenhan Yang, and Zongming Guo.
\newblock Context-aware text-based binary image stylization and synthesis.
\newblock {\em IEEE Transactions on Image Processing}, 28(2):952--964, 2018.

\bibitem{yang2018context1}
Shuai Yang, Jiaying Liu, Wenhan Yang, and Zongming Guo.
\newblock Context-aware unsupervised text stylization.
\newblock In {\em Proceedings of the 26th ACM international conference on
  Multimedia}, pages 1688--1696, 2018.

\bibitem{zhang2017synthesizing}
Junsong Zhang, Yu Wang, Weiyi Xiao, and Zhenshan Luo.
\newblock Synthesizing ornamental typefaces.
\newblock In {\em Computer Graphics Forum}, volume~36, pages 64--75. Wiley
  Online Library, 2017.

\bibitem{zhang2022creating}
Junsong Zhang, Zuyi Yang, Linchengyu Jin, Zhitang Lu, and Jinhui Yu.
\newblock Creating word paintings jointly considering semantics, attention, and
  aesthetics.
\newblock {\em ACM Transactions on Applied Perceptions (TAP)}, 19(3):1--21,
  2022.

\bibitem{zou2016legible}
Changqing Zou, Junjie Cao, Warunika Ranaweera, Ibraheem Alhashim, Ping Tan,
  Alla Sheffer, and Hao Zhang.
\newblock Legible compact calligrams.
\newblock {\em ACM Transactions on Graphics (TOG)}, 35(4):1--12, 2016.

\end{thebibliography}
}

\end{document}